\documentclass{article}

     \usepackage[preprint,nonatbib]{neurips_2020}

\usepackage[utf8]{inputenc} 
\usepackage[T1]{fontenc}    
\usepackage{hyperref}       
\usepackage{url}            
\usepackage{booktabs}       
\usepackage{amsfonts}       
\usepackage{nicefrac}       
\usepackage{microtype}      

\usepackage{graphicx}
\usepackage{booktabs}
\usepackage{array}
\usepackage{hhline}
\usepackage{upgreek}
\usepackage{comment}
\usepackage{color}
\usepackage{amsmath,amssymb}
\usepackage{ amssymb }
\usepackage{scalerel}

\newcommand{\reals}{\mathbb{R}}

\newcommand{\figref}[1]{Figure~\ref{#1}}

\newcommand{\secref}[1]{Section~\ref{#1}}

\renewcommand{\eqref}[1]{Eq.~\ref{#1}}

\definecolor{applegreen}{rgb}{0.55, 0.71, 0.0}

\title{Learning Interclass Relations for Image Classification}

\author{%
  Raouf ~Muhamedrahimov \quad Amir Bar \quad Ayelet Akselrod-Ballin \\
   Zebra Medical Vision Ltd. \\
   \texttt{\{raouf,amir,ayelet\}@zebra-med.com} \\
}

\begin{document}

\maketitle

\begin{abstract}


In standard classification, we typically treat class categories as independent of one-another. In many problems, however, we would be neglecting the natural relations that exist between categories, which are often dictated by an underlying biological or physical process. In this work, we propose novel formulations of the classification problem, based on a realization that the assumption of class-independence is a limiting factor that leads to the requirement of more training data. First, we propose manual ways to reduce our data needs by reintroducing knowledge about problem-specific interclass relations into the training process. Second, we propose a general approach to jointly learn categorical label representations that can implicitly encode natural interclass relations, alleviating the need for strong prior assumptions, which are not always available. We demonstrate this in the domain of medical images, where access to large amounts of labelled data is not trivial. Specifically, our experiments show the advantages of this approach in the classification of Intravenous Contrast enhancement phases in CT images, which encapsulate multiple interesting inter-class relations.

\end{abstract}


\section{Introduction}
In multi-class classification, class categories are typically treated as equally (or rather, infinitely) different from one another. In many tasks, however, this is not the case. In medical imaging, for example, discrete class categories often represent stages in continuous physiological processes, such as the progression of a pathology \cite{choi2018development,grassmann2018deep,kuang2019automated,peterfy2004whole}. Hence, in the standard representation of class categories, information on their underlying phenomena is lost.  Naturally, we would want to formulate our problem in a way that captures the ordinal nature of categories in the learning process to more closely reflect the reality of our data. Ordinal Regression \cite{armstrong1989ordinal, harrell2015regression, herbrich1999support} deals specifically these classification problems, where categories follow some natural order \cite{diaz2019soft,chu2007support,herbrich1999support,gu2014incremental}. In recent work, this type of classification has been performed by mapping discrete ground truth labels into a soft distribution during training, in order to incorporate ordinal relations \cite{diaz2019soft}. However, this approach requires prior knowledge or assumptions about the label domain and the precise ordinal nature of the categories. In this study, we extend on this framework and propose different approaches to encoding underlying ordinal relations into ground truth label representations, demonstrating that these can either be defined based on prior knowledge, or learnt from data. We argue that these approaches better represent the formulation of some problems, with the potential to improve the model performance.

The application domain we focus on is intravenous (IV) contrast enhancement phase classification in Abdominal CT. IV contrast administration and the resulting enhancement patterns are often critical in the diagnostic process in CT. Multiphase CT scans are acquired in distinct physiologic vascular time points after initial IV administration, such as the \textit{non-enhanced}, \textit{arterial}, \textit{venous} and \textit{delayed} phases \cite{choi2018development,bae2010intravenous,guite2013}. Information about the particular contrast phase of a CT scan relies upon manual entry by a technician and is often partially or inconsistently captured in the metadata associated with the scan (contained in the image format, DICOM). As such, an algorithmic solution to contrast phase classification is essential in permitting fully automated ML analysis of dynamic radiographic findings, capable of discerning, for example, between benign liver fibronodular hyperplasia and malignant hepatocellular carcinoma \cite{sun2017automatic,choi2018development}. 

This task is particularly interesting from an Ordinal Regression standpoint as its interclass similarities can be articulated in different ways. The categories (or phases) follow a temporal order -- the time following contrast injection. The visual features of the image, however, are cyclic -- following the delayed phase, the contrast agent is fully excreted and the scan returns to a non-enhanced appearance. Furthermore, the diagnostic characteristics overlap between categories such that a single clinical finding could potentially be diagnosed in multiple phases
\cite{bae2010intravenous,guite2013}.

In summary, we propose three formulations of the ordinal classification task, centered around introducing representations of interclass relations into the training process and demonstrate their effectiveness in the application of intravenous contrast phase classification in CT images. Our main contributions are as follows. (1) We demonstrate that encoding cyclic ordinal assumptions into ground-truth label representations during training improves classification performance over a naive one-hot approach in a medical imaging setting. We show that these improvements are most significant for small training sets, which are typical in the domain. (2) We propose two reformulations of the classification task in which label representations are learned from data. Under constraints, we demonstrate that natural ordinal relations can be implicitly learnt and encoded during training, leading to the same improvements in performance while requiring few prior assumptions. (3) To the best of our knowledge, this is the first time a circular ordinal regression approach has been employed in the medical imaging domain of IV contrast. Finally, while our experiments suggest these approaches may be particularly well suited to medical imaging tasks, the methods themselves are general to all domains and may lead to improvements outside the field of medical imaging.
 

\begin{figure}[t]
    \centering
    \includegraphics[width=\linewidth]{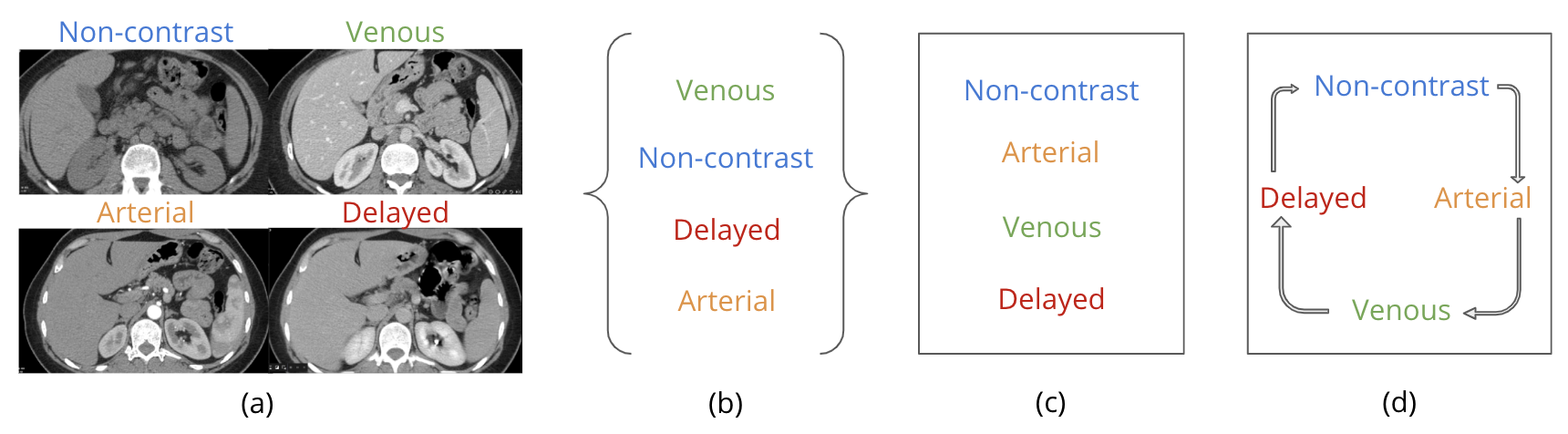}
    \caption{Examples of interclass relations for Intravenous Contrast flow phases. (a) An example of a CT abdomen image taken in four different phases. (b)-(d) demonstrate different class relations where in (b) class categories are unordered and in (c) contrast phases are ordered sequentially while in (d) phases are ordered in a circular manner. }
    \label{fig:teaser}
\end{figure}

\section{Related work}

\paragraph{Ordinal regression.} The goal of both classification and ordinal regression is to predict the category of an input instance $x \in \mathbb{R}^{H\times W\times D}$ from a set of $K$ possible class labels in set $\mathcal{Y}$. The difference is that in ordinal regression, there is a natural ordering (or ranking) associated with the classes \cite{diaz2019soft,chu2007support,herbrich1999support,gu2014incremental,cao2019rank, frank2001simple}. In some studies, an approach based on regression is taken; inputs are mapped to a continuous value in the label (or rank) space and $K-1$ thresholds are defined for categorization  \cite{herbrich1999support,chu2007support,gu2014incremental}. Other approaches formulate the task as a classification problem, treating thresholds in the rank-space as fixed and training classifiers over the $K$ ordinal categories \cite{diaz2019soft,cao2019rank, frank2001simple}. This study takes the latter approach and builds on recent work by \cite{diaz2019soft}, who propose the Soft Ordinal vectors (SORD) framework, where known ordinal information is encoded into the ground truth vectors through a soft target encoding scheme. Unlike previous work, we apply this in the medical domain, and our goal is to encode cyclic ordinal assumptions based on visual semantics. We extend their framework and propose an approach we term PL-SORD, where we allow the network to implicitly choose an optimal label encoding through reformulation of the training loss, defining a set of "candidates" for the categorical ordering. In some cases alleviates the need for strong prior assumptions with the same benefits to network supervision. 

\paragraph{Intravenous contrast} Recent works on IV contrast have modeled this as an image classification task, relying on human based annotations for training \cite{philbrick2018does,dercle2017impact,ma2020automated}. \cite{dercle2017impact,ma2020automated} propose systems to quality assess whether a scan
was accurately taken in Portal Venous Phase (PVP). The proposed approach in \cite{dercle2017impact} is semi-automatic and requires an expert in the loop. In \cite{ma2020automated}, a fully automatic system is proposed based on a CNN; however, the input is preprocessed such that it contains only a limited view of the image which includes the Aorta and Portal Vein. This constraint, might drop the overall algorithm
performance as described in \cite{dercle2017impact}. \cite{philbrick2018does} deal with contrast phase
classification with a view of a full abdominal CT slice, however, it is mainly used to examine neural networks visualization approaches in the context of clinical decision making. Moreover, this approach is based on a manually extracted 2D slices and human annotations, with a limited dataset of 3253 scans from a single institution. In this study, we also approach this as a classification task with the primary goal of demonstrating the effectiveness of introducing ordinal assumptions in label encodings. In terms of modelling, our network is based on a 3D representation of abdominal organs using training labels extracted from scan metadata. In doing so, we are able to build up a training set of 264,198 full CT scans from more than 10 institutions, supporting the generalizability of our experimental results.


\section{Overview of approaches}
Classification tasks performed over a set of discrete categories $\mathcal{Y}$, generally necessitate a label encoding $f:\mathcal{Y} \Rightarrow \mathbb{R}^{|\mathcal{Y}|}$, which maps any target class $t \in \mathcal{Y}$ (the ground-truth) to a vector of probability values $y_{\cdot|t} = f(t)$. The labels in this setting represent a probability distribution that the network attempts to match by optimizing for some loss metric $\mathcal{L}$ over a given training set. Naturally, any relation that might exist between a class $i \in \mathcal{Y}$ and the target class $t$ can be represented in the encoded value $y_{i|t} = f_i(t)$. 

\begin{figure}[t]
\centering
\includegraphics[width=\textwidth]{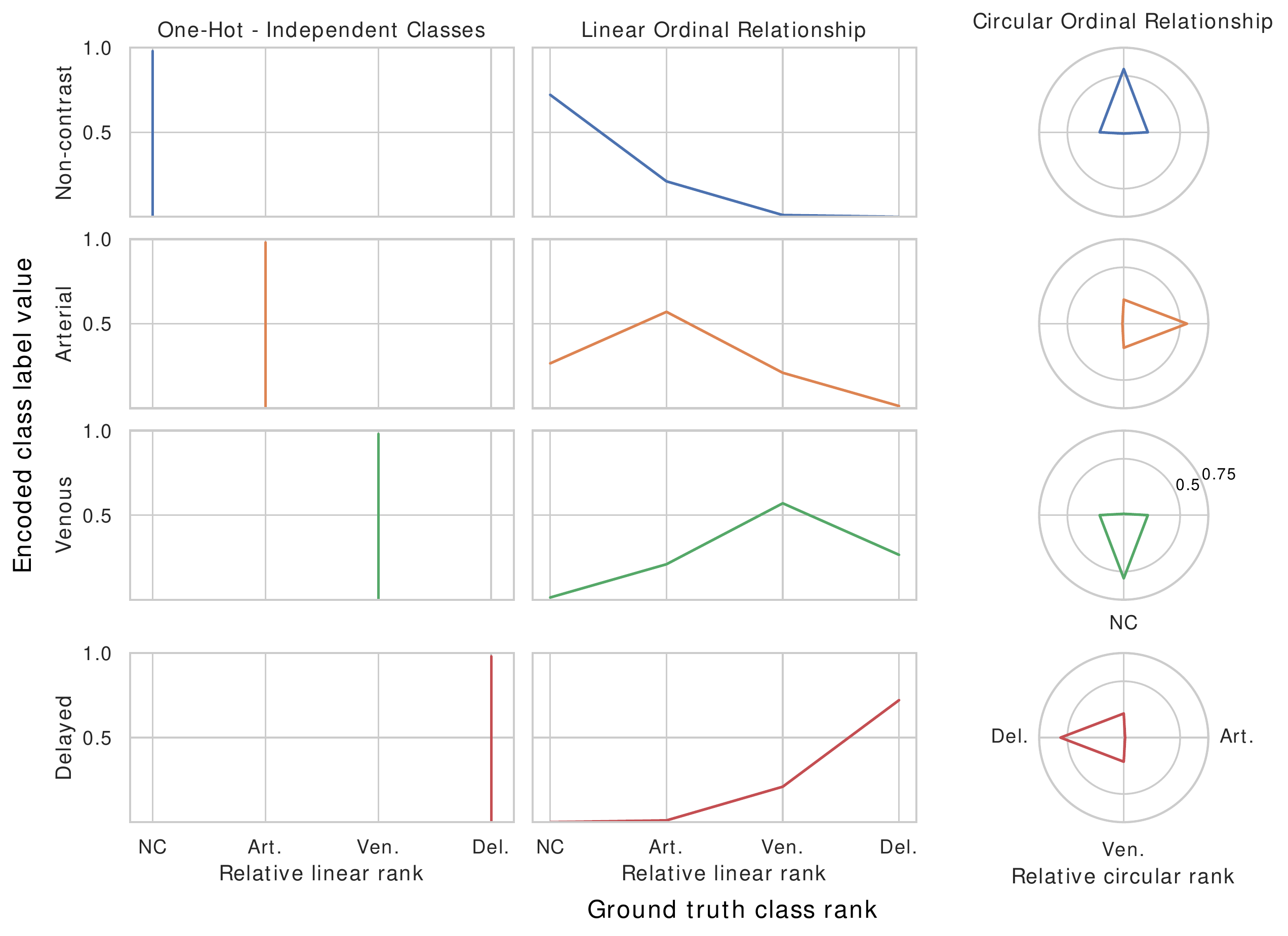}
\caption{Examples of SORD-encoded label representation values under different assumptions on the relations between classes: (left) no relations (one-hot encoding, $s=\infty$), (middle) linear relations, and (right) circular relations ($s=1.25\pi$). In each setting, categories considered adjacent to the true class are assigned a higher (non-zero) value than distant ones.} 
\label{fig:ordinal-labels}
\end{figure}

Classification tasks over $K$ classes are most commonly performed by representing each category as a one-hot vector (see \figref{fig:ordinal-labels}):
\begin{align}
y^{oh}_{i|t} = f_i(t) =  
    \begin{cases}
        1 \text{ if $i = t$} \\
        0 \text{ otherwise}
    \end{cases}
\label{eq:one-hot}
\end{align} 
In training, the network will treat all classes where $i \neq t$ as equally (or infinitely) wrong. In practice, this might not be the case -- some classes could be considered more correct than others. Biases may also exist in the data labels themselves. As the network tries match the label distribution exactly, there is motivation to define $f$ in a way that assigns a higher (non-zero) value to those categories closely related to a target class. 

In \secref{sec:sord}, we start by formulating the task as an ordinal regression problem by assigning $K$ ordinal positions (or ranks) corresponding to each class, $\Lambda = \{r_1, r_2, ..., r_K\}$ in some continuous domain $r_i \in \mathbb{R}^d$. Using the SORD encoding approach, we incorporate known class relations into label representations based on a pairwise interclass similarity metric $\phi(r_t, r_i)$, which represents the ordinal "distance" between the categories. Within this framework, we can represent the cyclic nature of the categories by simply ranking them in polar coordinates. Once the categorical ranks $\Lambda$ and distance metric $\phi$ for the task are specified based on prior knowledge, the predetermined mapping $f$ can be directly applied to the labels.

In \secref{sec:learnt-all}, instead of predefining a label encoding based on prior knowledge, we propose a more general approach whereby the label encoding $f$ is learnt from the data as part of the training process, exploiting ability of deep neural networks to generalize based on the visual semantics of the ordinal categories \cite{zhang2016understanding,maaten2008visualizing,chen2020simple,zeiler2014visualizing}.

In \secref{sec:circular-sord-perm}, we propose an approach we term PL-SORD, where we allow the network to implicitly choose an optimal label encoding $f$ defining a set of "candidates" $\Lambda \in S$ for the categorical ordering. 

Extending this, we loosen our constraints even further to see how this impacts the optimal label representation learnt from the data and the extent to which it reflects natural interclass similarities. Specifically, in \secref{sec:learn-cond}, we propose a formulation of the problem that attempts to directly learn a parametrized encoding function $f$ from the data, optimized jointly with the network parameters.




%
\section{Label encoding with known interclass relations}
\label{sec:sord}
We start by describing the formulation of a classification task as an ordinal regression problem, and define a fixed label representation $f$ using the SORD label encoding scheme based on prior assumptions on the ordinal relationship between categories. 

\subsection{Known categorical relations}
\label{sec:sord-base}
We define $K$ ordinal positions (or ranks) corresponding to each class, $\Lambda = \{r_1, r_2, ..., r_k\}$ in some continuous domain $r_i \in \mathbb{R}^d$. If the target class is $t$, the label representation for the class $i$ can be computed as the softmax over the pairwise interclass similarities:
 \begin{equation}
     y_{i|t} = f_i(t) = \frac{e^{-\phi(r_t,r_i)}}{\sum_{k=1}^{K}e^{-\phi(r_t,r_k)}}
\label{eq:ordinal-softmax}
\end{equation}
\noindent where $\phi(r_t, r_i)$ is a metric function representing the ordinal "distance" between the categories. Once the categorical ranks $\Lambda$ and distance metric $\phi$ for the task are specified, the mapping $f$ can be directly applied to the labels.


\subsection{Known circular relations}
\label{sec:circular-sord}
The gradual return to some original state can be accounted for by defining the ordinal categories as angles in polar coordinates (See \figref{fig:ordinal-labels}). The distance metric can then be defined as the shortest angular distance between two classes in this space:
\begin{align}
    \phi_{\theta_a}(r_t, r_i) &= \|r_i - r_t \mod 2\pi\|_1 \\
    \phi_{\theta_b}(r_t, r_i) &= 2\pi - \phi_{\theta_a}(r_t, r_i) \\
    \phi_{\theta}(r_t, r_i) &= (s \times \min(\phi_{\theta_a}, \phi_{\theta_b}))^2
\label{eq:phi-circular}
\end{align}
where $s$ can be understood as a scaling factor for the target probability distribution, treated as a network hyperparameter -- as $s \to \infty$, the label distribution becomes one-hot, as $s \to 0$, all $y_{\cdot|t}$ become equal.

\section{Learned label encoding}
\label{sec:learnt-all}
In this section, we propose a more generalized approach to jointly learn the label encoding $f$ to encapsulate categorical relations within the training process itself. 

\begin{figure}[t]
    \centering
    \includegraphics[width=1\textwidth]{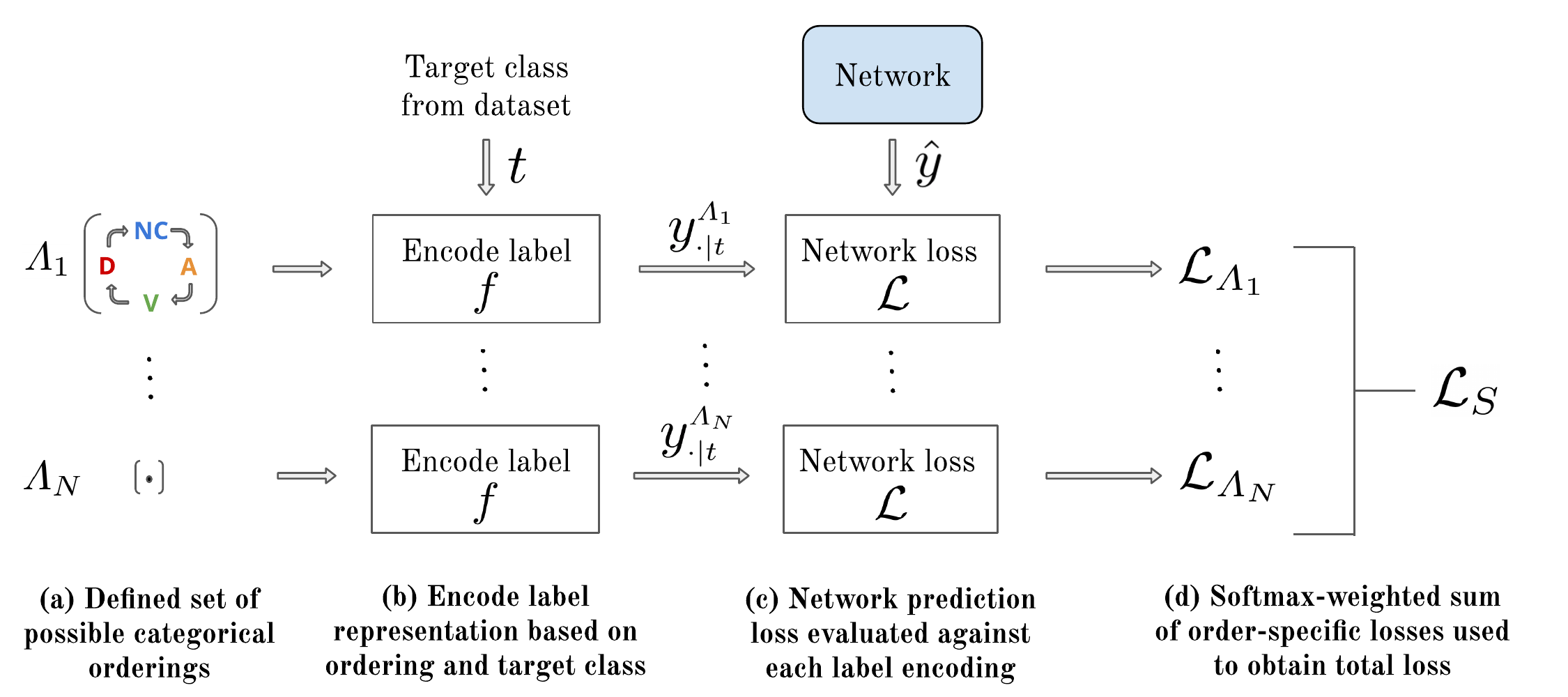}
    \caption{Summary of the PL-SORD approach. A set of proposed categorical orderings define separate label encodings, each used to calculate an order-specific loss for a single network prediction. During training, both the network parameters and the weighted contribution of each loss term to the total are optimized over the data.}
    \label{fig:pl-sord-diag}
\end{figure}

\subsection{Learning the optimal categorical ordering} 
\label{sec:circular-sord-perm}
Instead of pre-defining the categorical ordering $\Lambda$ as in (\secref{sec:circular-sord}), we relax this constraint and allow the network to choose from a set $S$ of possible categorical orderings (Figure~\ref{fig:pl-sord-diag}). We refer to this approach as PL-SORD. While the set of possible relations $S$ can be defined quite generally, here we propose one possible approach. We first specify the set of unique ordinal positions that can be assigned to any of our $K$ categories, so any rank $r_i$ must have a value from $\{R_1, ..., R_M\}$. The set $S$ of possible orderings is then taken as all size-$K$ permutations of these "available" positions $\{R_i\}_{i=1}^{M}$.

Using a fixed distance metric $\phi$ and \eqref{eq:ordinal-softmax}, each ordinal permutation $\Lambda \in S$ defines a separate label encoding $f(t; \Lambda) = y^{\Lambda}_{\cdot|t}$ as in \secref{sec:sord-base}, which can then used to calculate the order-specific loss for the prediction $\hat{y}$, without changes to the network:
\begin{align}
    \mathcal{L}_{\Lambda}(\hat{y}, y) &= \mathcal{L}(\hat{y}, y^{\Lambda})
\label{eq:perm-loss-single}
\end{align}

\noindent The total network loss is computed as a weighted sum of the $N$ order-specific losses for the same network, as illustrated in \figref{fig:pl-sord-diag}. So we optimize for both the loss weights $\lambda$ and the model parameters $W$ during training: 
\begin{align}
    \min_{W, \lambda}\mathcal{L}_S(\hat{y}, y) &= \sum_{j=1}^N\sigma_j(\lambda)\mathcal{L}_{{\Lambda_j}}
\label{eq:perm-loss-min}
\end{align}
Here, $\sigma_j(\lambda) = \frac{e^{\lambda_j}}{\sum_{k=1}^K e^{\lambda_k}}$ is the $j$th output of the softmax applied over the learned weights $\lambda \in \mathbb{R}^N$, and represents the contribution to the total loss from permutation $\Lambda_j$. Intuitively, the ordinal permutation which contributes the lowest training loss would be assigned the largest weight during training. 

\subsection{Directly learning label encoding}
\label{sec:learn-cond}

We alleviate the need to set constraints on distance metrics, rank-space, and symmetrical relationships between classes, and instead propose a formulation of the problem that attempts to directly learn the label encoding from the data. Specifically, when the target class is $t$, we hope to directly learn the distribution $y^{\alpha}_{i|t} = f(t, \alpha)$. It is clear that imposing no constraints would result in a degenerate solution, so we preserve the ground-truth signal by fixing the label value corresponding to the target class $t$ as $s \in (0, 1)$, which is treated as a hyperparameter, with the remaining $(1-s)$ distributed over the remaining classes. Concretely, the encoding for each target class $t$ is directly parametrized by $\alpha_t \in \reals^{K-1}$ and given by:
\begin{align}
y^{\alpha}_{i|t} = f_i(t; \alpha) &= 
    \begin{cases}
      s & \text{if $i = t$}\\
      (1-s)\frac{e^{{\alpha}_{t, i}}}{\sum_{k}e^{{\alpha}_{t, k}}} & \text{otherwise}
    \end{cases}     
\end{align}
During training, we optimize for both the network parameters $W$ and the encoding parameters $\alpha$:
\begin{align}
    \min_{W, \alpha}\mathcal{L}_S(\hat{y}, y) &= \mathcal{L}(\hat{y}, f(t; \alpha))
\label{eq:perm-loss-min}
\end{align}

\section{Experiments and results}

\subsection{Dataset}
For this study, we use a proprietary abdominal CT dataset consisting of 334,079 scans (181,881 patients) from over 10 institutions, with 90\%/10\%/10\% for training, validation, and test partitions, respectively. All 264,198 scans (144,144 patients) in the training partition were used for this study. For model validation, 1,000 scans (963 patients) were sampled with 250 scans from each class, equally sampled from 5 institutions. Four different labels were assigned to their respective contrast phases (\text{non-contrast}, \text{arterial}, \text{venous}, \text{delayed}). The labels were assigned using regular expressions applied to free text information contained in the SeriesDescription DICOM header. For testing, 192 CT scans were sampled from in the held-out test partition and were manually labeled by an expert radiologist with similar representation of each class, with 3 institutions represented within each class. Modelling approaches were compared on several randomly-sampled training sets with 2k, 4k, 8k, 16k, 32k, 80k, and 264k samples respectively.

All Patient Health Information (PHI) was removed from the data prior to acquisition, in compliance with HIPAA standards. The axial slices of all scans have an identical size of 512x512, but the number of slices in each scan vary between 42 and 1026, with slice spacing ranging from 0.45 mm to 5.0 mm. 

\subsection{Model and training}
The same model and training configuration was used for all experiments in this study. A region of interest, containing the liver, kidneys, aorta, and inferior vena cava (IVC), is automatically localized for each CT scan using an algorithmic approach similar to \cite{sahiner2018deep}. 20 input slices were uniformly sampled from the extracted region and each resized to $256 \times 256$. The input CT image pixel range was clipped in the range of a CT window centered at 40 with a width of 350 Hounsfield units.

We use a 3D ResNet50 architecture \cite{hara2018can,resnet} and apply a cross-entropy loss to the output of the model. The networks are trained on 2 NVIDIA GPUs using an Adam optimizer with a learning rate of $1\times10^{-4}$ and $\beta_1=0.9$, $\beta_2=0.999$ \cite{kingma2014adam}. We use a batch size of 32 with equal representation of samples from each class. We continue each training run for 150k iterations, measuring performance on the validation set every 150 iterations, saving the model with the highest categorical accuracy. Models were implemented using Keras (Tensorflow backend).

\subsection{Encoding known circular relationships}
Introducing prior knowledge about class relations, we compare the performance of SORD encoding approach described in \secref{sec:circular-sord} to a one-hot baseline on the task of contrast phase classification. We start by assigning equally-spaced ranks $r_i \in (0, 2\pi)$ to each of the categories in order of their physiological appearance: $r_{NC} = 0$, $r_{A} = \frac{\pi}{2}$, $r_{V} = \pi$, and $r_{D} = \frac{3\pi}{2}$ and assume a circular relationship exists between phases, in line with their visual and diagnostic features. Two scaling factors for the hyperparameter $s$ defined in \eqref{eq:phi-circular} are compared. Specifically, $s=0.625$, for which the loss is more centered around the ground truth class and $s=1$, where it is more distributed across adjacent classes.

\begin{figure}[t]
\centering
\includegraphics[width=0.6\textwidth]{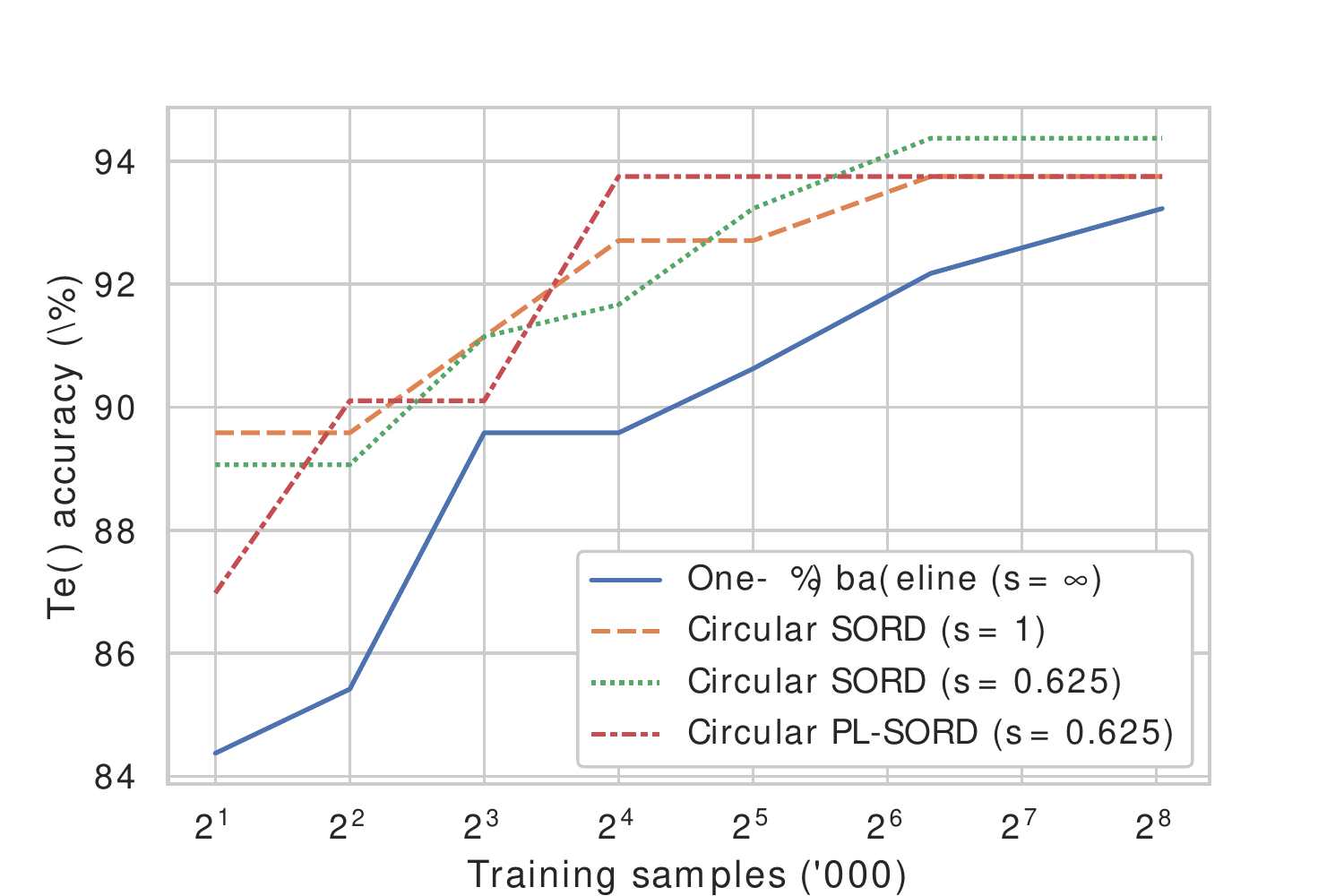}
\caption{Comparison of contrast phase classification test performance using, the one-hot encoding baseline ($s=\infty$), circular SORD encoding ($s=1$, $s=0.625$), and PL-SORD $s=1\pi$ with learnt ordinal ranks. The highest test accuracy for a training set equal or smaller in size is reported. Ordinal formulations result in higher accuracy across all training set sizes, most notably for small datasets.} \label{fig:results-core}
\end{figure}

In \figref{fig:results-core}, SORD encoding resulted in performance improvements across all training set sizes for both scaling factors $s=0.625$ and $s=1$. We see the most marked improvements for small datasets. Specifically, when holding everything else fixed, the models utilizing the Circular SORD achieve improved performance over the standard one-hot formulation even when presented less than $10\%$ of training data. 

%
\subsection{Learning the optimal ordinal encoding}
We compare the approach described in \secref{sec:circular-sord-perm} to a one-hot baseline and SORD encoding for contrast phase classification. A circular relationship is still assumed between phases and we allow them to take on 4 angular positions $r_i \in \{0, \tfrac{1}{2}\pi, \pi, \tfrac{3}{2}\pi\}$, without setting a constraint on the relative order. We then define our set of possible ordinal relations $S$ as all size-$K$ permutations of these equally-spaced positions. With an equal number of categories and possible positions, we are left with $|S| = \frac{(N-1)!}{2} = 3$ candidates for the natural ordering, after taking into account invariance to reversal and rotation in the circular setting\footnote{In the angular setting, the permutations $(0, \tfrac{1}{2}\pi, \pi, \tfrac{3}{2}\pi)$, $(\tfrac{3}{2}\pi, 0, \tfrac{1}{2}\pi, \pi)$, $(\tfrac{3}{2}\pi, \pi, \tfrac{1}{2}\pi, 0)$ are all equivalent}. Letting $\Lambda = (r_{NC}, r_{A}, r_{V}, r_{D}) \in S$ represent the assigned ranks of the non-enhanced, arterial, venous, and delayed phases, respectively, our set is given by:
\begin{align}
    S = \{\Lambda_1, \Lambda_2, \Lambda_3\} = \left\{ 
    (0, \tfrac{1}{2}\pi, \pi, \tfrac{3}{2}\pi), 
    (0, \pi, \tfrac{1}{2}\pi, \tfrac{3}{2}\pi), 
    (0, \tfrac{1}{2}\pi, \tfrac{3}{2}\pi, \pi) 
    \right\}
\end{align}
The approach was compared to the circular SORD and one-hot encoding for the same datasets and configuration. The results in \figref{fig:results-core} indicate that the performance is comparable to the Circular SORD, although it requires fewer assumptions. In a manual analysis of the learned weights ($\sigma(\lambda)$ in \eqref{eq:perm-loss-min}), it was found that the permutation that minimizes the network training loss follows the natural ordering of the classes, i.e. (``Non-Contrast", ``Arterial", ``Venous", ``Delayed"). So even if it seems that the ordering itself does not necessarily need to be enforced for an given task, representing the relationship in the labels can still be beneficial.
\subsection{Directly learning label representations}
\textbf{\begin{figure}[t]
\centering
\includegraphics[width=\textwidth]{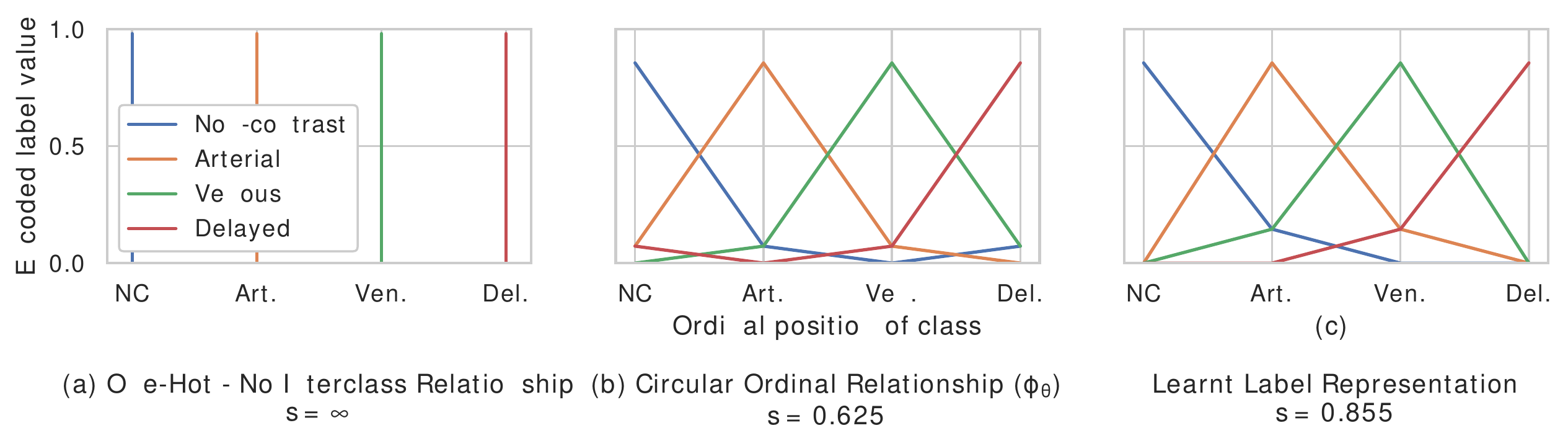}
\caption{Comparison of data label representations with a learnt encoding: (a) one-hot encoding baseline with unrelated classes ($s=\infty$), (b) circular relationship between classes ($s=1.25\pi$), and (c) learnt label representation.} \label{fig:learnt-labels}
\end{figure}}
The approach described in \secref{sec:learn-cond} of directly learning label representations is compared to the one-hot baseline on the contrast enhancement phase classification task. In this setting, the label encoding is learned from the data and no assumptions are imposed on the pairwise similarities; intuitively, encouraging the network to make certain mistakes above others. For comparison, we use the training set of 32k samples and set the hyper parameter $s=0.855$ (see \secref{sec:learn-cond}), such that we impose a maximum $y$ value which is similar to the circular SORD experiment. This model obtained a test accuracy of 92.28\% whereas the one-hot baseline achieved 90.63\%. 

We find the learnt label distribution $y^w_{i|t}$ (in \figref{fig:learnt-labels}c) results in an asymmetric encoding that could not be defined though the SORD framework described above, showing a higher degree of flexibility. However, it is worth noting that a "circular" relationship is not reflected, but rather weight is given to a single category responsible for the bulk of misclassifications for each target class. It is clear that the easiest-to-optimize encoding does not necessarily reflect the natural ordinal relations, without appropriate constraints on $f$. In additional to overlap of features, this highlights that some of the benefit of the SORD approach could be attributed to its accounting for biases, errors, or overlap in the labels of the data. 

While this does not result in improvements in performance over the SORD approaches, it offers a promising avenue for tasks where the relationship between categories can not be reasonably assumed, due to a large number of classes or limited knowledge of the problem. 


\section{Conclusion}
In this study we proposed three formulations of a classification task, centered around the introduction of interclass relations into label representations. In the application of intravenous contrast phase classification in CT images, we demonstrate that incorporating cyclic ordinal assumptions during training significantly improves classification performance over standard one-hot approaches, particularly when datasets are small. In the PL-SORD approach, we show that we can learn a label encoding that implicitly incorporates the natural ordinal relations, leading to the same improvements in performance while requiring fewer prior assumptions. Finally, by alleviating the need to set any ordinal assumptions and directly learning a label encoding from data, we again demonstrate improvements over a one-hot encoding. While the directly-learned approach does not result in accuracy improvements over approaches based in SORD (or exactly reflect our ordinal assumptions) it offers a promising avenue for tasks where the relationship between categories can not be reasonably assumed. It also highlights that some of the benefit of these approaches could be attributed to its accounting for biases, errors, or overlap in the labels of the data. All point to promising avenues for improving performance in medical classification tasks, where small datasets, ordinal relations, and noisy labels are exceedingly common.

\bibliographystyle{unsrt}
\bibliography{main.bib}

\end{document}